\documentclass[conference,10pt]{IEEEtran}
\IEEEoverridecommandlockouts
\usepackage{comment}
\usepackage[dvipsnames]{xcolor}
\usepackage{multirow}
\usepackage[ruled,linesnumbered]{algorithm2e}

\usepackage{cite}
\usepackage{amsmath,amssymb,amsfonts}
\usepackage{algorithmic}
\usepackage{graphicx}
\usepackage{textcomp}
\usepackage{xcolor}
\definecolor{darkgreen}{rgb}{0.0, 0.2, 0.13}
\def\BibTeX{{\rm B\kern-.05em{\sc i\kern-.025em b}\kern-.08em
    T\kern-.1667em\lower.7ex\hbox{E}\kern-.125emX}}
\begin{document}

\title{FF-INT8: Efficient Forward-Forward DNN Training on Edge Devices with INT8 Precision
\thanks{This work was supported in part by the National Science Foundation (Grant \#2350180, \#2453413, \#2312366, and \#2318152).}}

\IEEEaftertitletext{\vspace{-1.5\baselineskip}}

\author{\IEEEauthorblockN{Jingxiao Ma}
\IEEEauthorblockA{\textit{School of Engineering} \\
\textit{Brown University}\\
Providence, RI \\
jingxiao\_ma@alumni.brown.edu}
\and
\IEEEauthorblockN{Priyadarshini Panda}
\IEEEauthorblockA{\textit{Department of Electrical Engineering} \\
\textit{Yale University}\\
New Haven, CT \\
priya.panda@yale.edu}
\and
\IEEEauthorblockN{Sherief Reda}
\IEEEauthorblockA{\textit{School of Engineering} \\
\textit{Brown University}\\
Providence, RI \\
sherief\_reda@brown.edu}
}

\maketitle

\begin{abstract}
Backpropagation has been the cornerstone of neural network training for decades, yet its inefficiencies in time and energy consumption limit its suitability for resource-constrained edge devices. While low-precision neural network quantization has been extensively researched to speed up model inference, its application in training has been less explored.
Recently, the Forward-Forward (FF) algorithm has emerged as a promising alternative to backpropagation, replacing the backward pass with an additional forward pass. By avoiding the need to store intermediate activations for backpropagation, FF can reduce memory footprint, making it well-suited for embedded devices.
This paper presents an INT8 quantized training approach that leverages FF’s layer-by-layer strategy to stabilize gradient quantization. Furthermore, we propose a novel ``look-ahead'' scheme to address limitations of FF and improve model accuracy. Experiments conducted on NVIDIA Jetson Orin Nano board demonstrate 4.6\% faster training, 8.3\% energy savings, and 27.0\% reduction in memory usage, while maintaining competitive accuracy compared to the state-of-the-art. \end{abstract}


\section{Introduction}
\label{sec:intro}

Deep neural networks (DNNs) have achieved state-of-the-art performance across many application domains. However, their growing complexity significantly raises the computational cost of training, doubling roughly every six months from 2010 to 2022~\cite{sevilla2022compute}. 
For example, ResNet-50, which is introduced in 2015, consists of 26 million parameters and takes about 14 days to train on an NVIDIA M40 GPU~\cite{you2018imagenet}. In contrast, GPT-3, a large language model released in 2020, which contains 175 billion parameters, requires 3,640 PF-days to train, equivalent to 355 years of single-processor computing time, consuming 284,000 kWh of energy and costing more than 4.6 million US dollars~\cite{ouyang2022training}. 
These trends pose even greater challenges for edge devices, where limited memory and power resources make efficient, real-time DNN training an urgent priority.

The last few years have seen various methodologies for efficient deep learning computing, which can be categorized into pruning, quantization, neural architecture search, {\it etc}~\cite{abadade2023comprehensive}. Among these, quantization stands out for its clear impact on memory footprint and computational cost, achieved by reducing the precision of weights and activations without altering the number of learned features~\cite{ma2023wenet}. 
Moreover, many edge devices are equipped with specialized hardware optimized for low-precision computations, making quantization particularly well-suited for edge computing scenarios. 
Despite its benefits, most quantization techniques are limited to model inference~\cite{gholami2022survey}. Meanwhile, the Forward-Forward (FF) algorithm~\cite{hinton2022forward} offers an alternative to backpropagation, replacing the backward pass with an additional forward pass to address inefficiencies in gradient computation.
This paper delves into yet another advantage of the FF algorithm, specifically its applicability in implementing INT8 quantized training algorithms. The contributions of this paper are outlined below.

\begin{itemize}
    \item To the best of our knowledge, FF-INT8 is the first work to devise a low-precision training method leveraging the Forward-Forward algorithm. Our observation reveals that FF algorithm's layer-wise greedy approach effectively mitigates the accumulation of accuracy degradation, a common issue in quantized backpropagation techniques.

    \item We propose a novel training method based on the Forward-Forward algorithm, named FF-INT8. This approach adopts a greedy, layer-wise training strategy, where each layer of a deep neural network is trained independently using INT8 precision, making it especially well-suited for resource-constrained edge devices.

    \item We identify two key drawbacks of the Forward-Forward algorithm: low convergence accuracy and slow convergence speed. To address these limitations, we propose a novel ``look-ahead''  scheme that incorporates information from subsequent layers. This enhancement improves both training accuracy and convergence efficiency, particularly in low-precision training environments.

    \item Using FF-INT8, we train multiple DNNs on the Jetson Orin Nano board, a typical edge device. Our experimental results show that FF-INT8 significantly improves efficiency while preserving high accuracy. We achieve a speedup of 4.6\% in DNN training, reduce the memory footprint by 27.0\% and energy consumption by 8.3\% compared to a state-of-the-art INT8 training algorithm.
    
\end{itemize}

The organization of this paper is as follows. In section~\ref{sec:prev_work}, we overview relevant previous works in DNN quantization. We then briefly introduce Forward-Forward algorithm as background in section~\ref{sec:background}.  
In section~\ref{sec:methodology}, we introduce FF-INT8 training algorithm. We provide our experimental results in section~\ref{sec:results}. Finally, we summarize our conclusion in Section~\ref{sec:conclusion}.

\section{Previous Work}
\label{sec:prev_work}
A wide range of quantization methods have been proposed to enhance the efficiency of DNN inference, with post-training quantization (PTQ) being the most common approach.
PTQD~\cite{he2024ptqd} proposes a PTQ framework for diffusion models that optimizes model efficiency and performance by disentangling and correcting quantization noise.
To make the model more robust to the reduced precision, methodologies of quantization-aware training (QAT) are proposed to preserve end-to-end model accuracy post quantization~\cite{jacob2018quantization,chu2024make}. 

Compared to the large amount of studies on accelerating
inference by model quantization, few works explore low-precision training. MPTraining~\cite{micikevicius2018mixed} trains DNNs using 16-bit floating-point. Minifloat~\cite{fox2020block} also proposes to use 8-bit floating-point numbers to train DNNs. Floating-point quantization requires specific hardware platform to achieve acceleration, which is not conducive to the practical deployment on resource-constrained devices. On the other hand, compared to floating-point, quantizing into 8-bit integer has great potential in hardware acceleration, since INT8 operations are widely supported by recent GPUs. 
UI8~\cite{zhu2020towards} utilizes direction sensitive gradient clipping and deviation counteractive learning rate scaling to implement a fully unified INT8 training for convolutional neural networks. DAI8~\cite{zhao2021distribution} achieves better gradient quantization by performing channel-by-channel gradient distribution perception, but it sets too many channel dimension quantization parameters, adding additional computational complexity. GDAI8~\cite{wang2023gradient} also uses a data-aware dynamic quantization scheme to quantize various special gradient distributions. In summary, most INT8 training methods require extra effort to adapt to gradient distribution.

\section{Background}
\label{sec:background}

\begin{figure}[t!]
    \centering
    \includegraphics[width=0.85\linewidth]{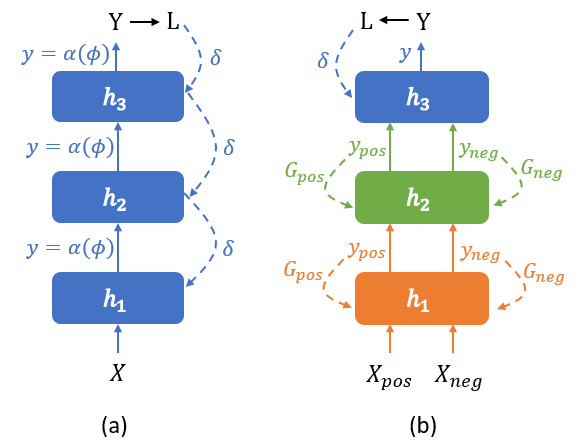}
    \vspace{-0.15in}
    \caption{(a) Backpropagation consists of a forward pass and a backward pass. (b) The Forward-Forward algorithm uses ``positive'' and ``negative'' datasets, and trains each layer individually using ``goodness'' function $G$. }
    \label{fig:BP_vs_FF}
    \vspace{-0.2in}
\end{figure}

The traditional backpropagation (BP) algorithm is inefficient in several ways. First, it requires substantial memory to store the full computational graph, leading to a large memory footprint. Additionally, the backward pass of derivative computation takes much longer time and consumes more energy.
In contrast, the Forward-Forward (FF) training algorithm~\cite{hinton2022forward} offers a novel direction to DNN training by replacing the forward and backward passes of BP with two forward passes to improve efficiency. Unlike BP, the FF algorithm uses two distinct data sets with opposing objective functions and trains the DNN layer-by-layer in a greedy manner. Figure~\ref{fig:BP_vs_FF} compares training procedure between BP and FF. In FF (Figure~\ref{fig:BP_vs_FF}b), the first step is to generate ``positive'' and ``negative'' datasets from input data. One way to generate datasets is to annotate the input data using one-hot encoding vector of the label~\cite{hinton2022forward}. More specifically, the one-hot vectors in positive samples are pointing to the true labels, while the ones in negative samples are pointing to the wrong labels. FF algorithm uses {\it goodness} function $G$ to qualify how well each layer is trained. The positive pass operates on positive samples and adjusts the weights to increase the goodness in
every hidden layer, while the negative pass operates on negative samples and adjusts the weights to decrease
the goodness. One common measurement of goodness function is the sum of squared neural activities. FF algorithm aims to make the goodness function well above a certain threshold $\theta$ for positive data ($G_{pos}$) and well below $\theta$ for negative data ($G_{neg}$), where $\theta$ can be considered as a hyperparameter that controls scale of the weights.

FF algorithm trains each layer once until convergence and then optimize its successor. The main idea behind such greedy scheme is to make each layer ``excited'' about positive
samples and, at the same time, less excited about negative ones, so that positive samples can be trained to match the correct label later.
FF algorithm improves efficiency in many aspects. First, instead of storing the entire computational graph, FF algorithm only stores current layer in memory, which significantly reduces memory footprint. Also, forward pass takes less time than backward pass. However, the heuristic property prevents earlier layers to learn from the later ones and final outputs. where training of earlier layers can be misleading. As a result, the FF algorithm often suffers from accuracy loss and struggles to scale effectively to modern, large-scale DNNs.

\section{Proposed Methodology}
\label{sec:methodology}
In this section, we describe our proposed methodology of INT8 Forward-Forward algorithm for efficient DNN training. Our FF-INT8 training method builds upon symmetric uniform quantization (SUQ), which is one of the most efficient quantization methods due to its hardware-friendly computation~\cite{gholami2022survey}.

\subsection{Network Depth and Gradient Quantization}
\label{sec:depth}

\begin{figure}[t!]
    \centering
    \includegraphics[width=0.94\linewidth]{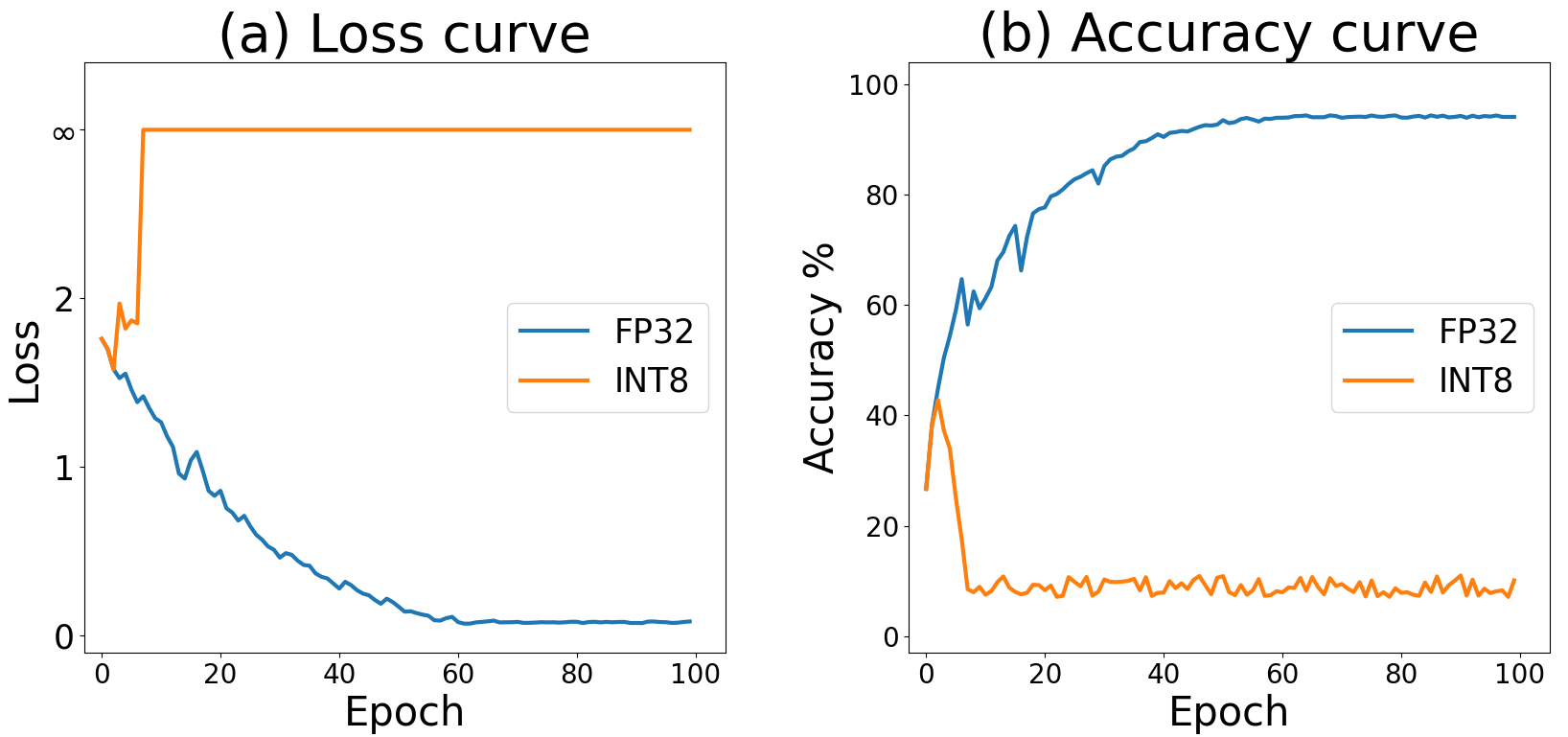}
    \vspace{-0.12in}
    \caption{Loss and accuracy of ResNet-18 on CIFAR-10 when gradients are directly quantized to INT8.}
    \label{fig:direct_quantize}
    \vspace{-0.15in}
\end{figure}

\begin{figure}[t!]
    \centering
    \includegraphics[width=\linewidth]{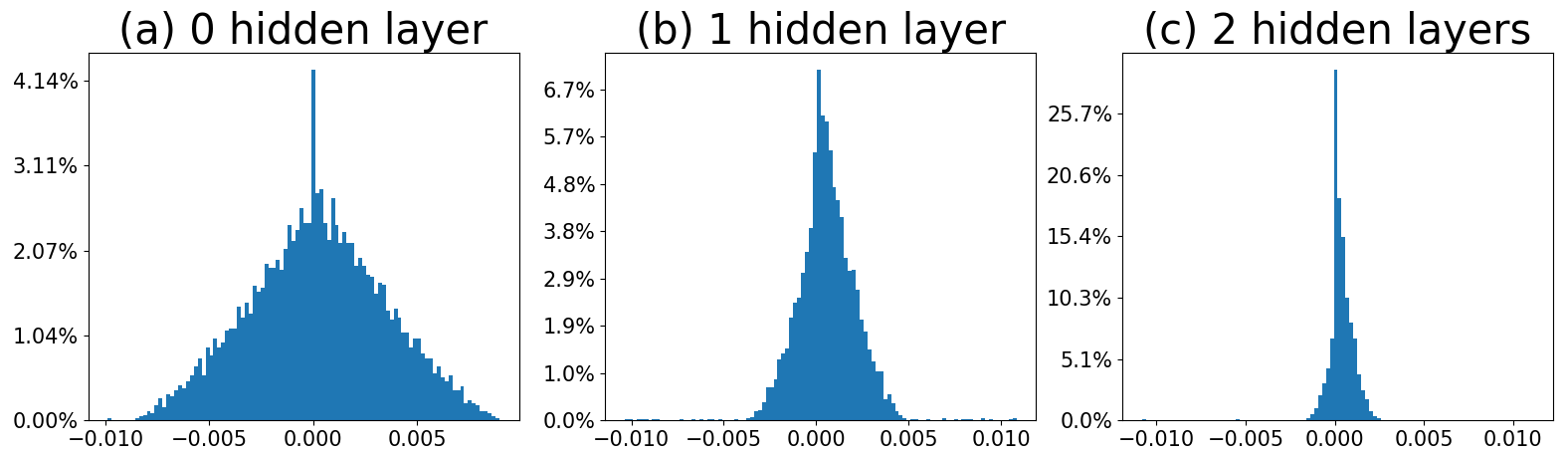}
    \vspace{-0.25in}
    \caption{Gradient distribution of first layer with different num. of hidden layers.}
    \label{fig:gradient_distribution}
    \vspace{-0.15in}
\end{figure}

As a preliminary experiment, we train ResNet-18 using CIFAR-10 dataset, with backpropagation and INT8 quantization including gradients. Figure~\ref{fig:direct_quantize} compares the changes of loss and accuracy per epoch between 32-bit floating-point (FP32) and INT8 training. The network is trained properly with FP32 precision. However, the loss of INT8 training increases dramatically as soon as we start training, while the accuracy drops to random level. We assume that INT8 training fails due to the accumulation of quantization error through backward propagation of derivatives. To further prove this hypothesis, we train three multilayer perceptrons (MLP) with different number of layers on MNIST dataset with different number of hidden layers, each of which is trained with FP32 and INT8 precision respectively. As Table~\ref{table:net_depth} shows, as the number of hidden layers increases, the accuracy of FP32 training increases until the model overfits. On the contrary, the accuracy of INT8 training decreases dramatically as the network becomes deeper. The accuracy difference between FP32 and INT8 training is considerably small with a single-layer network, but increases significantly as we include the first hidden layer. We may conclude that quantization error accumulates as network becomes deeper. Instead of deeper networks, INT8 training can be directly executed on a single-layer network. Finally, we plot gradient distribution of the {\it first} layers for each MLP in Table~\ref{table:net_depth}, which are trained using FP32. As Figure~\ref{fig:gradient_distribution} shows, for deeper networks, gradient distribution of earlier layers are sharper with larger extreme values, while distribution of single-layer network is more even. Thus, direct quantization in deeper network leads to large quantization error, since most gradients gather in a small range and we cannot tell the differences in such small range after quantization. On the other hand, direct quantization in single-layer network is less error-prone. FF algorithm proposes to train each layer individually in a greedy manner, making it particularly compatible with INT8 training.

\begin{table}[t!]
  \centering
    \caption{Accuracy of multilayer perceptrons on MNIST dataset with different number of hidden layers and training precision. Each hidden layer consists of 500 neurons.}
     \vspace{-0.05in}
  \begin{tabular}{|c|c|c|c|}
\hline
Number of & FP32 & INT8 & Accuracy\\
Hidden Layers & Acc. (\%) & Acc. (\%) & Difference (\%) \\
\hline
0 & 89.5 & 88.7 & -0.8 \\
1 & 93.4 & 73.8 & -19.6 \\
2 & 94.5 & 62.4 & -32.1 \\
3 & 94.3 & 65.2 & -29.1 \\\hline
  \end{tabular}
  \label{table:net_depth}
  \vspace{-0.05in}
\end{table}

\subsection{INT8 Forward-Forward Algorithm}
\label{sec:FF-INT8}

\begin{figure}[t!]
    \centering
    \includegraphics[width=0.93\linewidth]{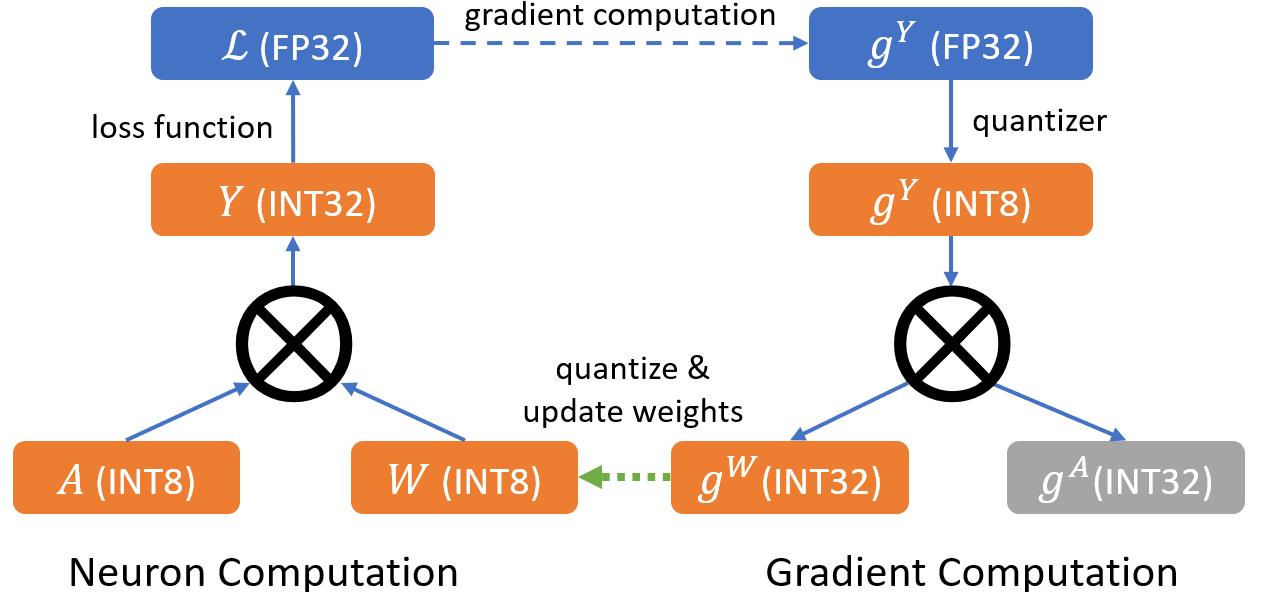}
    \vspace{-0.1in}
    \caption{Dataflow of FF-INT8 on a single layer.}
    \label{fig:method}
    \vspace{-0.15in}
\end{figure}

FF-INT8 algorithm trains each layer individually in a greedy manner using both positive and negative datasets, with only one layer being trained at a time. For each dataset (positive or negative), our FF-INT8 training workflow for a single layer is illustrated in Figure~\ref{fig:method}. Initially, we apply SUQ with stochastic rounding~\cite{gupta2015deep} to quantize the input data $A$. INT8 MAC operations are then employed to compute the dot product in either dense or convolution layers. Specifically, integer matrix multiplication with INT8 inputs and INT32 accumulation is used. The loss function is based on the negative log-likelihood of the goodness function $G$. For the positive dataset, the loss function $\mathcal{L}_{pos}$ is computed as follows:
\vspace{-0.07in}
\begin{equation}
    \mathcal{L}_{pos} = -\log p(positive) = \log{(1+ e^{-(G_{pos} -\theta)})}
    \label{eq:loss_pos}
\vspace{-0.07in}
\end{equation}
For negative dataset, loss function $\mathcal{L}_{neg}$ is computed as
\vspace{-0.07in}
\begin{equation}
      \mathcal{L}_{neg} = -\log p(negative) = \log{(1+ e^{(G_{neg} -\theta)})}
      \label{eq:loss_neg}
    \vspace{-0.07in}
\end{equation}
where $G$ represents the goodness function, and $\theta$ is the hyperparameter that represents threshold as discussed in Section~\ref{sec:background}. 

By optimizing this loss function, we encourage outputs of positive samples to be large and outputs of negative samples to be small. The gradient $g^Y$ is computed and quantized into INT8. Since gradients are not back-propagated to previous layers, there is no need to compute the gradients of input data $g^A$. Only gradients of weights $g^W$ are computed and updated to current weights. Thus, for computation of neuron activity $Y$ and gradient $g_W$, we replace floating-point computations with INT8, which saves a large amount of computation.

\subsection{FF-INT8 Algorithm with ``Look-Ahead''}
\label{sec:multi-round}

As discussed in Section~\ref{sec:background}, FF algorithm offers efficiency but imposes limitations by restricting each layer to learn solely from its immediate input, preventing communication with subsequent layers, particularly the final outputs. Consequently, earlier layers are primarily trained to distinguish between positive and negative data, leaving the task of specific label classification to the final layer. This lack of feedback from final outputs can lead to suboptimal training of earlier layers. As a result, FF algorithm often suffers from lower accuracy and requires more epochs to converge compared to backpropagation. These challenges are exacerbated in deeper networks, rendering the FF algorithm less effective for training large-scale DNNs. Additionally, FF algorithm struggles to accommodate widely-used structures such as residual blocks, which allow inputs to bypass certain layers and contribute directly to the block's output~\cite{he2016deep}. To optimize residual blocks effectively, all layers within the block must be trained interactively. However, due to the layer-by-layer nature of FF algorithm, layers within the same block are trained independently, without awareness of subsequent layers, leading to significant accuracy loss.

\begin{algorithm}[t!]
	\small
	\SetKwBlock{Begin}{begin}{end}
	\SetKwInOut{Input}{Input}
	\SetKwInOut{Output}{Output}
	\Input{Training set $(\textbf{X}, \textbf{Y})$, Number of epochs $n$}\label{alg:input}
	\Output{Trained network $M$ with $k$ layers}
	Initialize neural network $M$ \\
        Initialize hyperparameter $\lambda=0$ in loss function \\
        Generate positive and negative sets from $(\textbf{X}, \textbf{Y})$ \\
        \For{$num\_epoch = 1,2,\ldots, n$} {
        Execute forward pass using INT8 precision\\
        Compute goodness function $G= ||\mathbf{y} ||^2$ for each layer\\
        \For{current layer $= 1,2,\ldots, k$}{
        Compute loss for current layer using $G$ \\
        Update weights using gradients in INT8 precision\\
        }
        Increase hyperparameters $\lambda$ in loss function\\
        }
	\caption{FF-INT8 with ``Look-Ahead''}
		\label{alg:look_ahead}

\end{algorithm}

\begin{figure}[t!]
\vspace{-0.15in}
    \centering
    \includegraphics[width=0.75\linewidth]{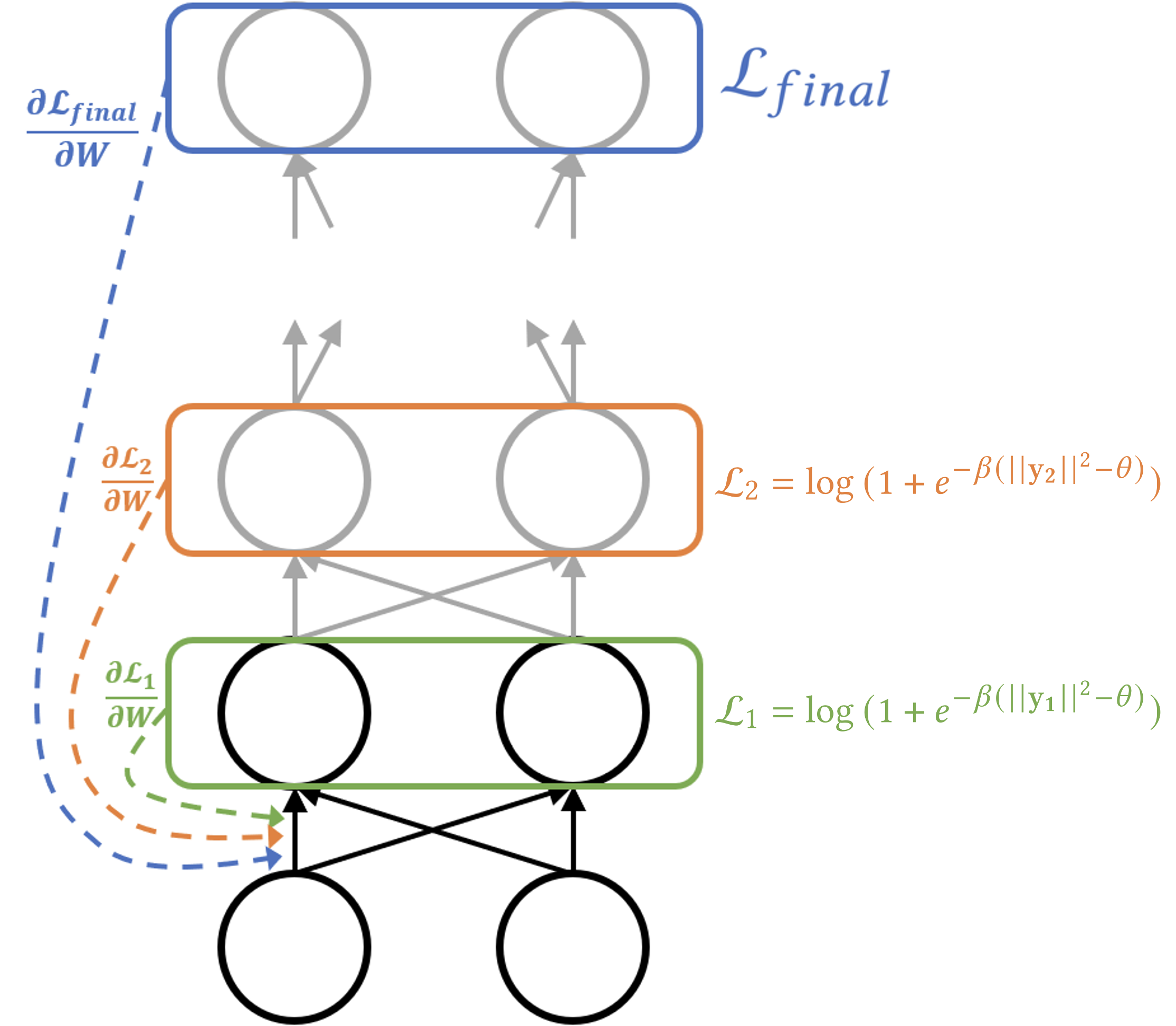}
    \vspace{-0.1in}
    \caption{Gradient computation of FF algorithm with ``look-ahead'', where loss functions of later layers are considered.}
    \label{fig:updated}
    \vspace{-0.15in}
\end{figure}

To solve these issues, our solution is to bridge connections between different layers, especially with final outputs, while retaining the efficiency of the FF algorithm. We name our method ``{\it Look-Ahead}'' scheme. In the standard FF algorithm, the loss function considers only the goodness function of the current layer during training. To enable feedback from subsequent layers, we redefine the loss function to incorporate the goodness functions of these layers. The revised loss function is defined as:
\vspace{-0.05in}
\begin{equation}
    \mathcal{L}_{new} = \mathcal{L}_{1} + 
    \lambda \times (\mathcal{L}_{2} + \mathcal{L}_{3} + \cdots + \mathcal{L}_{final})
    \label{eq:update}
    \vspace{-0.05in}
\end{equation}
Here, $\mathcal{L}_{i}$ represents the loss of the $i^{th}$ layer, derived from the goodness function, and $\lambda$ is a coefficient that balances the contributions of the current layer and the subsequent layers. Figure~\ref{fig:updated} illustrates the derivative computation with this modified loss function, as follow:
\vspace{-0.03in}
\begin{equation}
    \frac{\partial  \mathcal{L}_{new} }{\partial W} = \frac{\partial  \mathcal{L}_{1} }{\partial W} + \lambda\times\frac{\partial  (\mathcal{L}_{2} + \mathcal{L}_{3} + \cdots + \mathcal{L}_{final}) }{\partial W}
    \label{eq:gradient}
    \vspace{-0.01in}
\end{equation}

In Section~\ref{sec:FF-INT8}, we described a methodology where each layer is trained for $n$ epochs independently, without considering information from later layers. For a neural network with $k$ layers, this approach requires a total of $k \times n$ derivative computations. With the new loss function introduced in Equation~\ref{eq:update}, which involves derivatives from subsequent layers, we must perform complete forward passes for each derivative calculation. For the sake of efficiency, we propose Algorithm~\ref{alg:look_ahead} to utilize {\it one} forward pass to update all layers. Specifically, for each epoch, we execute complete forward pass (line 5), and compute goodness function $G$ for each layer (line 6). We can then construct loss functions of all layers as Equation~\ref{eq:update} using goodness function of each layer (line 8), and compute the gradient (line 9). This approach facilitates interaction between layers, enabling them to optimize simultaneously and improve accuracy. Importantly, the goodness function for each layer depends solely on its neuron values, and the loss function is based solely on the goodness function. As a result, constructing the backward derivative chain is unnecessary, maintaining the computational efficiency of the loss and derivative calculations as vanilla FF-INT8 algorithm.
Overall, the total number of derivative computations remains $k\times n$, the same as before. As later demonstrated in Section~\ref{sec:modify_results}, the ``look-ahead'' scheme significantly reduces the number of epochs needed for convergence, further improving efficiency. Regarding memory usage, since this scheme requires a full forward pass to compute the goodness function of later layers, all network weights must be retained in memory, resulting in a modest increase in memory footprint. However, compared to backpropagation, this approach still saves memory by avoiding the storage of intermediate activations and the full gradient chain, as it does not rely on a complete backward pass. 

One more hyperparameter in ``FF-INT8 Algorithm with look-ahead'' is $\lambda$, which balances between current layer and other layers. For first few epochs, since each layer is less optimized, our priority is to train each layer so that it can basically tell between positive samples and negative samples. In other word, we want to discourage the interaction between layers at the beginning of training, since later layers are not optimized and information of later layers does not help the optimization of current layer.
After a few epochs, each layer is better optimized. We then gradually increase $\lambda$ to encourage interaction between layers to improve convergence accuracy.

\section{Experimental Results}
\label{sec:results}

In this section, we first introduce our experimental setup. The first experiment is to compare FF-INT8 algorithm with and without ``look-ahead'', to show the benefit of it. We then theoretically analyze and compare the number of each operation, demonstrating that our methodology reduces computational cost. Finally, we comprehensively use our methodology to train different DNNs, and compare against other methodologies to demonstrate state-of-the-art performance.

\vspace{-0.05in}
\subsection{Experimental Setup}
\label{sec:setup}
\begin{enumerate}
    \item Datasets and models: As shown in Table~\ref{table:data_model}, we evaluate four different DNNs as benchmarks.  Since the experiments are conducted on an edge device with limited computation resources, we mainly use CIFAR10 as training dataset. The batch size is set to 32.
    \item Hardware setup: The NVIDIA Jetson Orin Nano board, equipped with an INT8 engine, is used to measure training time, energy and memory usage. The device specifications, detailed in Table~\ref{table:spec}, demonstrate that it is representative of edge devices used for DNN training.
    \item Hyperparameters: In Equation~\ref{eq:loss_pos} and~\ref{eq:loss_neg}, we set threshold $\theta = 2.0$. In Equation~\ref{eq:update}, $\lambda$ is initialized to 0, and increased by 0.001 each epoch.
\end{enumerate}

\begin{table}[t!]
  \centering
    \caption{DNN Architectures and Datasets}
        \vspace{-0.05in}
  \begin{tabular}{|c|c|c|}
\hline
\multirow{2}{*}{DNN} & \multirow{2}{*}{Dataset} & Num. of \\ 
& & Params (M) \\\hline\hline
Multi-layer perceptron (MLP)& \multirow{2}{*}{MNIST} & \multirow{2}{*}{1.79} \\ 
(2 hidden layers)&  &  \\ 
\hline
MobileNet-V2~\cite{sandler2018mobilenetv2} & \multirow{3}{*}{CIFAR10} & 2.24 \\ \cline{1-1}\cline{3-3}
EfficientNet-B0~\cite{tan2019efficientnet} &  & 3.39 \\ \cline{1-1}\cline{3-3}
ResNet-18~\cite{he2016deep} &  & 11.19 \\ \hline

  \end{tabular}
  \label{table:data_model}
  \vspace{-0.1in}
\end{table}

\begin{table}[t!]
  \centering
    \caption{Technical Specifications of NVIDIA Jetson Orin Nano}
        \vspace{-0.05in}
  \begin{tabular}{|l|l|}

\hline
GPU & 512-core NVIDIA Ampere architecture GPU \\ \hline
CPU & 6-core Arm\textregistered    
Cortex\textregistered-A78AE v8.2 64-bit CPU \\ \hline
Memory & 4GB 64-bit LPDDR5 34 GB/s \\  \hline
Power & 7W-10W \\ 
\hline
  AI Performance & 20 TOPS \\
\hline

  \end{tabular}
  \label{table:spec}
\end{table}

\begin{figure}[t!]
    \centering
    \includegraphics[width=\linewidth]{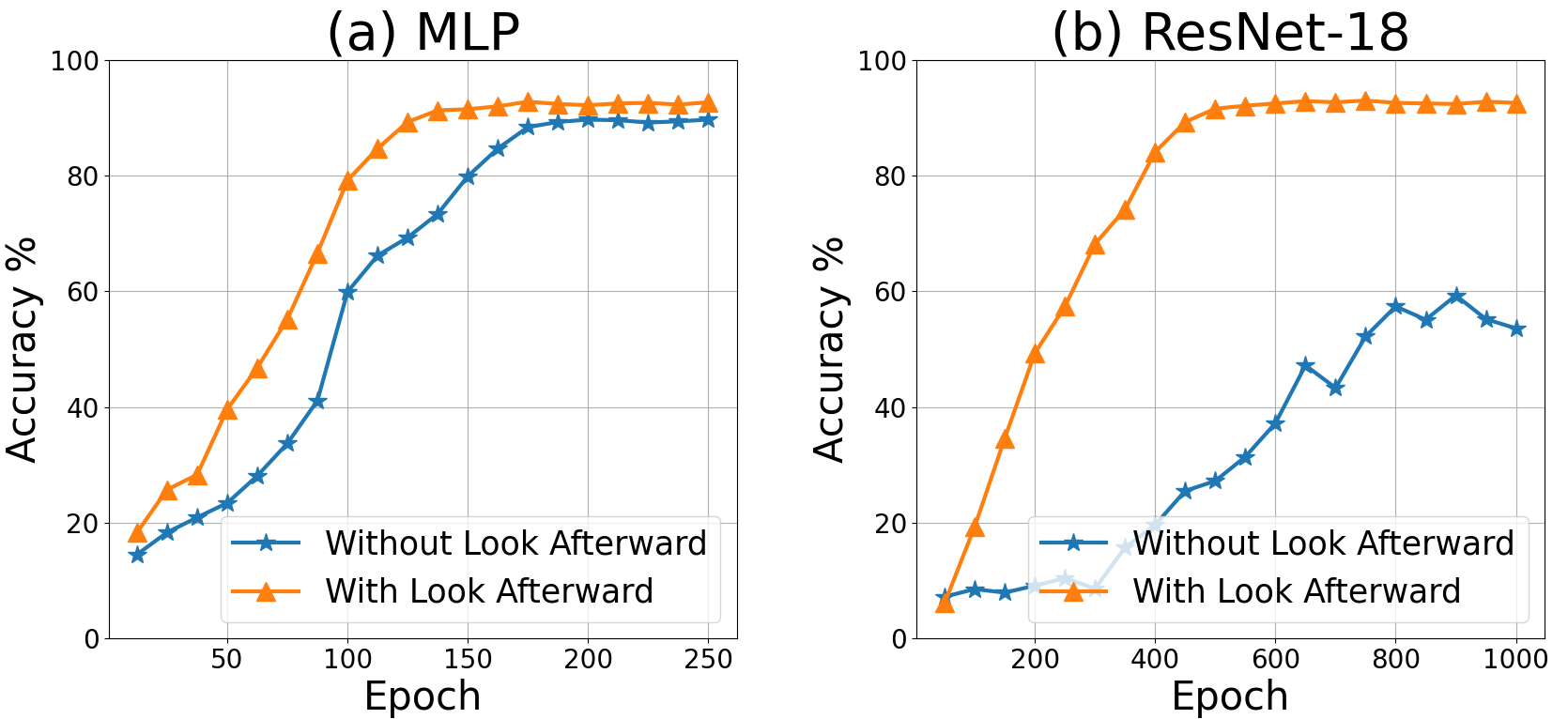}
    \vspace{-0.2in}
    \caption{Test accuracy across different epochs for MLP and ResNet-18 trained using FF-INT8, with and without ``look-ahead'' scheme respectively.}
    \label{fig:num_rounds}
    \vspace{-0.15in}
\end{figure}

\subsection{Training with ``Look-Ahead''}
\label{sec:modify_results}

In the first set of experiment, we show that ``look-ahead'' scheme improves accuracy for FF-INT8 as described in Section~\ref{sec:multi-round}. We first train MLP with 2 hidden layers using both FF-INT8 algorithms and demonstrate results in Figure~\ref{fig:num_rounds}(a). Without ``look-ahead'' scheme, the accuracy converges to nearly 90\% with 180 epochs. However, with ``look-ahead'' scheme, MLP reaches slightly higher convergence accuracy with only 130 epochs. Thus, we argue that by interaction with later layers, earlier layers compute gradients and update weights in a ``more optimized way'', which leads to higher convergence accuracy and faster training.

Figure~\ref{fig:num_rounds}(b) demonstrates the results in ResNet-18, which consists of residue blocks. Different from MLP, if ResNet-18 is trained without ``look-ahead'' scheme, the accuracy converges to only 60\%, while the accuracy curve is very unstable. There are two reasons for the bad performance:
\begin{enumerate}
    \item The scale of ResNet-18 is much larger compared to MLP, which means it is much difficult for a greedy approach like FF algorithm to find the optimized solution. 
    \item Residue block allows input to be skipped over few layers and accumulated to block's output. However, without ``look-ahead'' scheme, when previous layers in a block are trained, the vanilla FF algorithm does not consider the accumulation operation in block's output layer. And when block's output layer is trained, the previous layers have already trained and cannot be modified. Thus, the optimization process is limited, which leads to low convergence accuracy.
\end{enumerate}
As Figure~\ref{fig:num_rounds}(b) shows, by considering subsequent layers, ``look-ahead'' scheme solves both problems and significantly improves convergence accuracy. Given that many modern DNN architectures incorporate residual blocks,  such method significantly enhances the FF-INT8 algorithm.

\begin{table}[t!]
  \centering
    \caption{Comparison of computational cost between INT8 Forward-Forward algorithm and INT8/FP32 backpropagation}
        \vspace{-0.1in}
  \begin{tabular}{|c|c|c|c|}
\hline
\multicolumn{2}{|c|}{Computation} & Operation & Counts (OPs) \\ \hline \hline
\multirow{4}{*}{FF-INT8} & Quantization & 32-bit CMP & 32.4K \\ \cline{3-4}
& Phase & 32-bit FADD & 165.9K \\ \cline{2-4}
& MAC & 8-bit MUL & 23.8M \\ \cline{3-4}
& Phase & 8-bit ADD & 23.8M \\ \hline \hline
\multirow{2}{*}{BP-FP32} & MAC & 32-bit FADD & 898.2M \\ \cline{2-4}
& Phase & 32-bit FMUL & 898.2M  \\ \hline 
\hline
 & Quantization & 32-bit CMP & 7.2K \\ \cline{3-4}
GDAI8~\cite{wang2023gradient}& Phase & 32-bit FADD & 18.4K \\ \cline{2-4}
(BP-INT8)& MAC & 8-bit MUL & 898.2M \\ \cline{3-4}
& Phase & 8-bit ADD & 898.2M \\ \hline 
\hline

  \end{tabular}
  \label{table:comp_cost}
  \vspace{-0.2in}
\end{table}

\begin{table*}[t!]
  \centering
    \caption{Summary of model accuracy, training time, energy consumption and memory footprint between different approaches. Training algorithms are based on either BP (backpropagation) or FF (the Forward-Forward algorithm). The suffix denotes the precision, where FP32 means 32-bit floating-point, and INT8 means quantizing to 8-bit integer. UI8~\cite{zhu2020towards} refers to unified INT8 training algorithm, and GDAI8~\cite{wang2023gradient} refers to gradient distribution-aware INT8 training algorithm.}
        \vspace{-0.1in}
  \begin{tabular}{|l||l|l|l|l|l|}
\hline

Model & Training Algorithm & Accuracy (\%)&  Time (s)&  Energy (J) &  Memory (MB) \\ \hline \hline

\multirow{5}{*}{MLP} & \textcolor{red}{BP-FP32} & \textcolor{red}{94.5} & \textcolor{red}{482.3} & \textcolor{red}{2315.0}  & \textcolor{red}{247.6} \\ 
& BP-INT8 & 52.4 & 326.1 & 1206.6 & 213.9 \\ 
& BP-UI8~\cite{zhu2020towards} & 92.3 & 335.2 & 1277.1 & 197.0 \\ 
& \textcolor{blue}{BP-GDAI8~\cite{wang2023gradient}} & \textcolor{blue}{93.8} & \textcolor{blue}{344.9} & \textcolor{blue}{1345.4} & \textcolor{blue}{182.6} \\ 
& \textcolor{darkgreen}{FF-INT8} &
 \textcolor{darkgreen}{94.3} \textcolor{red}{{\scriptsize (-0.2)}} \textcolor{blue}{{\scriptsize (+0.5)}}&
 \textcolor{darkgreen}{312.7} \textcolor{red}{{\scriptsize (-35.2\%)}} \textcolor{blue}{{\scriptsize (-9.3\%)}}&
 \textcolor{darkgreen}{1097.0} \textcolor{red}{{\scriptsize (-52.6\%)}} \textcolor{blue}{{\scriptsize (-18.4\%)}}& 
 \textcolor{darkgreen}{140.7}\textcolor{red}{{\scriptsize (-43.2\%)}} \textcolor{blue}{{\scriptsize (-22.9\%)}} \\ \hline \hline

\multirow{5}{*}{MobileNet-v2} & \textcolor{red}{BP-FP32} & \textcolor{red}{91.5} & \textcolor{red}{2370.8} & \textcolor{red}{11593.2} & \textcolor{red}{649.8} \\ 
& BP-INT8 & 5.9 & 1851.6 & 7836.0 & 571.6 \\ 
& BP-UI8~\cite{zhu2020towards} & 87.2 & 1960.0 & 7618.5 & 592.6 \\ 
& \textcolor{blue}{BP-GDAI8~\cite{wang2023gradient}} & \textcolor{blue}{90.9} & \textcolor{blue}{1790.7} & \textcolor{blue}{6528.1} & \textcolor{blue}{578.9} \\ 
& \textcolor{darkgreen}{FF-INT8} &
 \textcolor{darkgreen}{91.1} \textcolor{red}{{\scriptsize (-0.4)}} \textcolor{blue}{{\scriptsize (+0.2)}}&
 \textcolor{darkgreen}{1703.9} \textcolor{red}{{\scriptsize (-28.1\%)}} \textcolor{blue}{{\scriptsize (-4.8\%)}}&
 \textcolor{darkgreen}{6174.3} \textcolor{red}{{\scriptsize (-46.7\%)}} \textcolor{blue}{{\scriptsize (-5.4\%)}}& 
 \textcolor{darkgreen}{437.0} \textcolor{red}{{\scriptsize (-32.7\%)}} \textcolor{blue}{{\scriptsize (-24.5\%)}} \\ \hline \hline

\multirow{5}{*}{EfficientNet-B0} & \textcolor{red}{BP-FP32} & \textcolor{red}{89.4} & \textcolor{red}{2692.8} & \textcolor{red}{13356.2} & \textcolor{red}{861.0} \\ 
& BP-INT8 & 11.8 & 2095.0 & 8563.9 & 703.9 \\ 
& BP-UI8~\cite{zhu2020towards} & 85.3 & 2230.8 & 8656.2 & 735.5 \\ 
& \textcolor{blue}{BP-GDAI8~\cite{wang2023gradient}} & \textcolor{blue}{88.9} & \textcolor{blue}{2177.1} & \textcolor{blue}{8589.9} & \textcolor{blue}{692.0} \\ 
& \textcolor{darkgreen}{FF-INT8} &
 \textcolor{darkgreen}{ 88.6} \textcolor{red}{{\scriptsize (-0.8)}} \textcolor{blue}{{\scriptsize (-0.2)}}&
 \textcolor{darkgreen}{ 2129.9} \textcolor{red}{{\scriptsize (-20.9\%)}} \textcolor{blue}{{\scriptsize (-2.2\%)}}&
 \textcolor{darkgreen}{ 8093.8} \textcolor{red}{{\scriptsize (-39.4\%)}} \textcolor{blue}{{\scriptsize (-5.8\%)}}& 
 \textcolor{darkgreen}{ 505.2} \textcolor{red}{{\scriptsize (-41.3\%)}} \textcolor{blue}{{\scriptsize (-27.0\%)}} \\ \hline \hline

\multirow{5}{*}{ResNet-18} & \textcolor{red}{BP-FP32} & \textcolor{red}{93.5} & \textcolor{red}{3853.0} & \textcolor{red}{18764.1} & \textcolor{red}{1096.4} \\ 
& BP-INT8 & 7.2 & 2676.1 & 10436.8 & 885.8 \\ 
& BP-UI8~\cite{zhu2020towards} & 89.7 & 2873.8 & 11466.5 & 920.7  \\ 
& \textcolor{blue}{BP-GDAI8~\cite{wang2023gradient}} & \textcolor{blue}{92.9}  & \textcolor{blue}{2751.6} & \textcolor{blue}{10291.0} & \textcolor{blue}{894.1} \\ 
& \textcolor{darkgreen}{FF-INT8} &
 \textcolor{darkgreen}{93.1} \textcolor{red}{{\scriptsize (-0.4)}} \textcolor{blue}{{\scriptsize (+0.2)}}&
 \textcolor{darkgreen}{ 2697.9} \textcolor{red}{{\scriptsize (-30.0\%)}} \textcolor{blue}{{\scriptsize (-2.0\%)}}&
 \textcolor{darkgreen}{ 9926.5} \textcolor{red}{{\scriptsize (-47.1\%)}} \textcolor{blue}{{\scriptsize (-3.5\%)}}& 
 \textcolor{darkgreen}{ 682.3} \textcolor{red}{{\scriptsize (-37.7\%)}} \textcolor{blue}{{\scriptsize (-23.7\%)}} \\ \hline \hline

 \multicolumn{2}{|l|}{\textbf{Avg. difference between FF-INT8 }}& \multirow{2}{*}{ \textbf{\normalsize Reduce 0.4\%} } & \multirow{2}{*}{ \textbf{\normalsize Save 28.6\%}} & \multirow{2}{*}{ \textbf{\normalsize Save 46.4\%}} & \multirow{2}{*}{ \textbf{\normalsize Save 38.7\%} }\\
 \multicolumn{2}{|l|}{\textbf{and BP-FP32 (baseline) }} &&&&\\ \hline

 \multicolumn{2}{|l|}{\textbf{Avg. difference between FF-INT8 }}& \multirow{2}{*}{ \textbf{\normalsize Improve 0.2\%} } & \multirow{2}{*}{ \textbf{\normalsize Save 4.6\%}} & \multirow{2}{*}{ \textbf{\normalsize Save 8.3\%}} & \multirow{2}{*}{ \textbf{\normalsize Save 27.0\%} }\\
 \multicolumn{2}{|l|}{\textbf{and BP-GDAI8 (state-of-the-art) }} &&&&\\ \hline

  \end{tabular}
  \label{table:comp_results}
  \vspace{-0.2in}
\end{table*}

\subsection{Analysis of Computational Cost}
\label{sec:analysis}

In Section~\ref{sec:modify_results}, our experiment suggests that FF-INT8 training requires a large number of epochs to converge. In this section, we analyze the theoretical computational cost to demonstrate that although FF-INT8 training has much more epochs compared to backpropagation, it is still efficient. We count the number of required operations to train 4-layer MLP using MNIST dataset with three settings, {\it i.e.} the proposed FF-INT8 training method with ``look-ahead'', and backpropagation with 32-bit floating-point (BP-FP32) as baseline. We also include Gradient Distribution-Aware INT8 training algorithm (GDAI8)~\cite{wang2023gradient} for comparison, which is an INT8 training algorithm based on backpropagation. Table~\ref{table:comp_cost} summarize the amount of computation required for training a mini-batch of 10 samples for three approaches. In FF-INT8, we have a quantization phase and a multiply–accumulation (MAC) phase, where computation of quantization phase is negligible compared to MAC. Comparing MAC phase, FF-INT8 training requires 23.8M 8-bit MAC operations, whereas backpropagation approach (BP-FP32 or GDAI8) requires 898.2M MAC operations (FP32 or INT8). This is because the FF algorithm does not have large matrix multiplication to back-propagate derivatives from the last layer to the first. Thus, per mini-batch, FF-INT8 only requires $2.6\%$ of MAC operations in backpropagation approach. INT8 arithmetic is also 4x faster than FP32 in hardware, which makes FF-INT8 hundres of times more efficient compared to BP-FP32 per epoch. Additionally, FF-INT8 only computes forward pass and does not compute backward pass of gradients. In many devices, forward pass is more efficient due to hardware optimization for model inference.
In conclusion, although FF-INT8 training needs a large number of epochs, theoretically it is still more efficient compared to both backpropagation approaches.

\subsection{Accuracy, Time, Energy, Memory Footprint}

For the last set of experiment, we compare model accuracy, training time, energy consumption between multiple training algorithms, and demonstrate results in Table~\ref{table:comp_results}. As {\it baseline}, we train DNNs using backpropagation (BP) with 32-bit floating-point operations, denoted as \textcolor{red}{{BP-FP32}}. If directly quantizing gradients to INT8 using BP (BP-INT8), although we observe significant time, energy and memory savings, accuracy decreases dramatically due to the accumulation of quantization error, as analyzed in Section~\ref{sec:depth}. With deeper architecture, direct quantization makes training even worse.

As a comparison, we train each DNN with two existing INT8 training algorithms, {\it i.e.} Unified INT8 Training Algorithm (BP-UI8)~\cite{zhu2020towards} and Gradient Distribution-Aware INT8 Training Algorithm (\textcolor{blue}{BP-GDAI8})~\cite{wang2023gradient}. Both algorithms are based on BP, and quantize gradients based on analysis of gradient distribution. Since they use BP with additional operation of gradient monitoring and analysis, the computational cost of both is slightly higher than direct quantization (BP-INT8). The training accuracy is close to traditional BP with FP32 operands (baseline). \textcolor{blue}{BP-GDAI8} has the {\it state-of-the-art} performance in terms of trade-off between model accuracy and efficiency, which is chosen 
as the comparison model.

Finally, we implement our FF-INT8 with ``look-ahead'' as described in Section~\ref{sec:multi-round}. As analysis in Section~\ref{sec:FF-INT8}, gradient quantization works better on FF algorithm due to its layer-by-layer greedy manner. Thus, the training accuracy is much higher compared to direct quantization in BP (BP-INT8). By comparing to BP-GDAI8, we show that training accuracy of FF-INT8 is 0.2\% higher against the state-of-the-art, and is very close to the traditional backpropation (BP-FP32), with a 0.4\% reduction.
For training time and energy consumption, since Jetson Orin Nano board has an INT8 engine, we observe a significant improvement compared to training with FP32 operations. Our FF-INT8 algorithm is also more efficient compared to BP-GDAI8, with 4.6\% saving in training time and 8.3\% saving in energy consumption. 

As mentioned in Section~\ref{sec:background}, one of the major benefits of FF algorithm is smaller memory footprint. 
Normally, BP algorithms rely on automatic differentiation, which requires to store the large computational graph for gradient backpropagation in memory. But since FF algorithm does not have backward pass, it does not need to store this computational graph. Such improvement in memory footprint is further enhanced by INT8 operations. Thus, compared to BP-GDAI8, FF-INT8 saves memory footprint by 27.0\%.

\section{Conclusion}
\label{sec:conclusion}

In this paper, we present FF-INT8, a novel low-precision training methodology that leverages the Forward-Forward algorithm for INT8 quantized training. Unlike backpropagation-based approaches, FF-INT8 stabilizes gradient quantization by employing a layer-wise greedy training strategy, effectively mitigating accuracy degradation. To further enhance convergence and accuracy, we propose the ``look-ahead'' scheme, which enables earlier layers to incorporate feedback from subsequent layers, addressing a key limitation of the standard FF algorithm.  By comparing against the state-of-the-art approach, we demonstrate that FF-INT8 significantly reduces training time, energy cost and memory footprint while maintaining high accuracy. FF-INT8 offers a promising alternative for energy-efficient neural network training.

\bibliography{ref}

\end{document}